\documentclass[runningheads]{llncs}

\usepackage[year=2026,ID=*****]{eccv}
\usepackage{eccvabbrv}
\usepackage{graphicx}
\usepackage{booktabs}
\usepackage[accsupp]{axessibility}
\usepackage{hyperref}
\usepackage{orcidlink}
\usepackage[capitalize]{cleveref}
\usepackage{siunitx}

\begin{document}

\title{Modality Discrepancy Transformer for Ambivalence and Hesitancy Recognition}

\titlerunning{Modality Discrepancy Transformer for A/H Recognition}

\author{Shiyu Luo\inst{3,2}\and
Yu Wang\inst{1,2}\and
Jiawen Huang\inst{1,2}\and
Zhaoxiang Xiao\inst{4,2}\and
\mbox{Chenxi Huang}\inst{2,1}\and
Qi Zhang\inst{1,2}\and
Bin Liu\inst{2,1}}

\authorrunning{S.~Luo et al.}

\institute{School of Artificial Intelligence, University of Chinese Academy of Sciences \and Institute of Automation, Chinese Academy of Sciences \and School of future technology, University of Chinese Academy of Sciences \and
College Of Computer and Information Engineering, Tianjin Normal University \\
\email{luoshiyu221@mails.ucas.ac.cn}, \email{wangyu230@mails.ucas.ac.cn},
\email{huangjiawen25@mails.ucas.ac.cn}, \email{xuegaodef@163.com},
\email{huangchenqian22@mails.ucas.ac.cn}, \email{zhangqi2025@ia.ac.cn},
\email{liubin@nlpr.ia.ac.cn}}

\maketitle

\begin{abstract}
  Ambivalence and hesitancy (A/H) are affective states in which individuals
  express contradictory signals across facial, vocal, and linguistic channels.
  Automatically recognising A/H in clinical videos requires detecting cross-modal
  disagreement---the signal that standard fusion methods suppress. Based on
  the conflict-aware multimodal fusion framework of Bekhouche~\etal,
  we present the Modality Discrepancy Transformer (MDT). MDT enriches the
  original 6-token design to a 9-token representation comprising three modality
  embeddings, three absolute-difference features, and three Hadamard-product
  discrepancy features learned through linear projections. These nine tokens
  undergo Transformer self-attention, with FiLM-based text-conditioned modulation
  and LoRA fine-tuning as core architectural components. A text-guided late fusion
  branch blends a text-only auxiliary head with the full multimodal output at
  inference. On the BAH dataset from the 3rd ABAW Challenge, MDT achieves
  \textbf{0.7408} Macro F1 on the labelled test split and \textbf{0.7368} on the
  private leaderboard, outperforming the strongest published baseline by over 10
  points while training in under 20 minutes on a single GPU.
\end{abstract}

\section{Introduction}
\label{sec:intro}

Ambivalence is the simultaneous experience of conflicting attitudes; hesitancy
is the reluctance to commit. Both are central to behavioural medicine: a patient
may verbally agree to treatment while their face and voice express
doubt. Detecting such mismatches supports
motivational interviewing and addresses challenges including vaccine
hesitancy. A/H differs fundamentally from basic emotion
recognition: happiness manifests consistently across channels, whereas A/H is
defined by cross-modal contradiction---saying one thing while showing another.
Standard multimodal fusion is designed to amplify inter-modal agreement; it can
actively suppress the disagreement signals that characterise A/H.

The BAH dataset~\cite{gonzalez2026bah} (1,427 videos, 300 participants, 778
training samples, participant-wise splits) provides the benchmark for this problem
through the ABAW A/H Challenge~\cite{kollias2026abaw10}. Prior work established
a clear modality hierarchy: Savchenko~\cite{savchenko2024abaw8} achieved strong
results with text-dominant features, and the BAH annotator analysis confirms that
face--language inconsistency accounts for 42.6\% of A/H
cues~\cite{gonzalez2026bah}. However, text-only approaches systematically
over-detect A/H because hedging language appears in clinical speech even without
genuine ambivalence, producing a large gap between positive- and negative-class
performance.

Bekhouche~\etal~\cite{bekhouche2026conflictaware} introduced a conflict-aware
architecture (CA-AH) that computes pairwise absolute differences between
modality embeddings, yielding a 6-token representation fed to a Transformer.
This approach established explicit discrepancy modelling for A/H, achieving
0.715 Macro F1 on the ABAW10 leaderboard. We extend this framework in three
directions. First, the discrepancy representation is enriched from three
absolute-difference features to six features: adding Hadamard-product
discrepancy vectors that capture per-dimension interaction patterns through
learned projections which yields a 9-token Transformer input. Second,
text-conditioned FiLM modulation~\cite{perez2018film} of video and audio
features enables the linguistic channel to directly shape how visual and
acoustic information is expressed. Third, LoRA~\cite{hu2022lora} replaces layer
unfreezing, substantially reducing trainable parameters and overfitting risk on
the 778-example dataset.

Our contributions are: (1)~a 9-token enriched discrepancy representation that
combines absolute-difference and Hadamard-product features to provide a more
complete description of cross-modal relationships; (2)~the integration of FiLM
modulation and LoRA fine-tuning as core components rather than optional
enhancements; (3)~a comprehensive ablation study demonstrating that MDT achieves
0.7408 Macro F1 on the labelled test split and 0.7368 on the ABAW11 private
leaderboard, outperforming all published baselines by over 10 points.

\section{Related Work}
\label{sec:related}

\subsection{A/H Recognition and the ABAW Challenges}
The ABAW series~\cite{kollias2024abaw,kollias2025abaw8,kollias2026abaw10} has been instrumental in advancing multimodal affective computing, providing a standardised benchmark for in-the-wild emotion recognition. Within the A/H track, Gonz'{a}lez \emph{et al.}~\cite{gonzalez2026bah} established baseline results on the BAH dataset, reporting video-level scores of 0.593 (LFAN~\cite{zhang2023lfan}) and 0.634 (zero-shot M-LLM). In contrast, Savchenko~\cite{savchenko2024abaw8} demonstrated that frame-level approaches with text-dominant features achieve superior performance (0.772 Macro F1), highlighting the significance of lexical content. More recently, Bekhouche \emph{et al.}~\cite{bekhouche2026conflictaware} proposed the CA-AH Transformer, a 6-token architecture specifically designed for conflict-aware fusion, which attained 0.715 on the ABAW10 leaderboard. Building upon CA-AH, our MDT introduces richer discrepancy representation and cross-modal modulation, combined with parameter-efficient fine-tuning, to better handle the limited data regime of the BAH dataset.

\subsection{Multimodal Encoders}
For unimodal representation, we leverage well-established pretrained backbones: VideoMAE-Base~\cite{tong2022videomae} (86M) for visual feature extraction, HuBERT-Base~\cite{hsu2021hubert} (94M) for audio, and RoBERTa-GoEmotions~\cite{demszky2020goemotions,lowe2022roberta,liu2019roberta} (125M) for text. Notably, unlike window-based text alignment, we adopt the full transcript as input, motivated by the video-level nature of A/H recognition, which benefits from global contextual understanding.

\subsection{Multimodal Fusion and Small-Data Learning}
Multimodal fusion can be broadly classified into early, late, and hybrid approaches~\cite{baltruvsaitis2019multimodal}, with recent work emphasising the importance of modelling inter-modal discrepancies. While disagreement-aware mechanisms have shown promise in sentiment~\cite{poria2017sentiment} and deception analysis~\cite{perez2015deception}, their application to the 778-example BAH dataset remains challenging due to severe overfitting risks. To address this, our framework integrates LoRA~\cite{hu2022lora} for parameter-efficient updates, Focal Loss~\cite{lin2017focal} to mitigate class imbalance, and CutMix~\cite{yun2019cutmix} for feature-space regularisation, ensuring stable training under limited annotations.

\section{Method}
\label{sec:method}

\begin{figure}[tb]
  \centering
  \includegraphics[width=\linewidth]{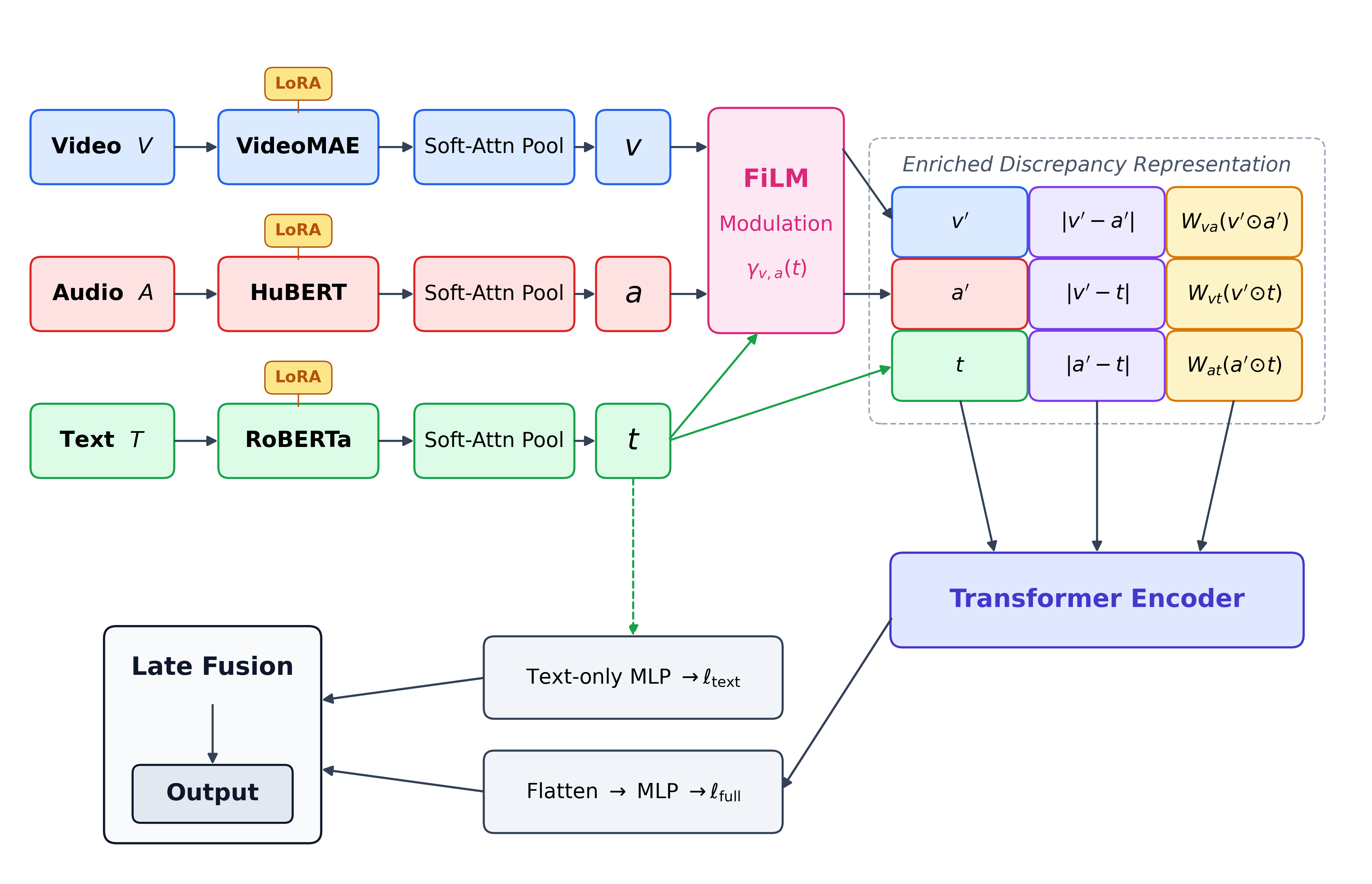}
  \caption{MDT overview. Three frozen encoders produce modality embeddings,
    pooled via soft-attention and projected to $\mathbf{v},\mathbf{a},\mathbf{t}$.
    Six discrepancy features (three absolute-difference, three Hadamard-product)
    form a 9-token sequence processed by a 2-layer Transformer. Text-conditioned
    FiLM modulates video and audio before discrepancy computation. LoRA adapters
    fine-tune each encoder. A text-only auxiliary head blends with the multimodal
    output at inference.}
  \label{fig:architecture}
\end{figure}

MDT predicts a binary label $y\in\{0,1\}$ from video $V$, audio $A$, and
transcript $T$, optimising Macro F1 per the challenge
protocol~\cite{gonzalez2026bah}. \Cref{fig:architecture} gives an overview.

\subsection{Modality Encoding}

VideoMAE-Base processes $T=16$ face frames $(224\!\times\!224)$; HuBERT-Base
encodes 16\,kHz mono audio with padding masking; RoBERTa-GoEmotions encodes the
full Whisper~\cite{radford2023whisper} transcript. Each encoder output is projected
to $D=768$ and aggregated via modality-specific learnable soft-attention
pooling~\cite{bahdanau2015attention}:

\begin{equation}
  \mathbf{e} = \sum_i \alpha_i \mathbf{x}_i, \qquad
  \alpha_i = \frac{\exp(\mathbf{w}^\top \mathbf{x}_i)}{\sum_j \exp(\mathbf{w}^\top \mathbf{x}_j)},
  \label{eq:attn}
\end{equation}

Padding positions are masked to $-\infty$ before the softmax so that they receive
zero weight. The resulting pooled embeddings are
$\mathbf{v},\mathbf{a},\mathbf{t}\in\mathbb{R}^{D}$.

\subsection{Enriched Discrepancy Representation}
\label{sec:discrepancy}

The central premise is that A/H manifests as cross-modal inconsistency:
face--language disagreement is the dominant cue (42.6\% of annotated
segments~\cite{gonzalez2026bah}). CA-AH~\cite{bekhouche2026conflictaware}
used three absolute-difference features: $|\mathbf{v}-\mathbf{a}|$,
$|\mathbf{v}-\mathbf{t}|$, $|\mathbf{a}-\mathbf{t}|$. These capture the
magnitude of divergence per dimension but provide no information about
how modalities interact within each dimension. To capture both aspects,
MDT augments these with three Hadamard-product discrepancy features:

\begin{equation}
  \begin{aligned}
    \mathbf{d}_{va} &= |\mathbf{v} - \mathbf{a}|, \;\;
    \mathbf{d}_{vt} = |\mathbf{v} - \mathbf{t}|, \;\;
    \mathbf{d}_{at} = |\mathbf{a} - \mathbf{t}|, \\[2pt]
    \mathbf{h}_{va} &= \mathbf{W}_{va}(\mathbf{v} \odot \mathbf{a}), \;\;
    \mathbf{h}_{vt} = \mathbf{W}_{vt}(\mathbf{v} \odot \mathbf{t}), \;\;
    \mathbf{h}_{at} = \mathbf{W}_{at}(\mathbf{a} \odot \mathbf{t}),
  \end{aligned}
  \label{eq:discrepancy}
\end{equation}

where $\odot$ denotes element-wise multiplication and $\mathbf{W}_{*}\in\mathbb{R}^{D\times D}$
are learned projection matrices. The Hadamard product captures per-dimension
interaction between modality pairs; the projection maps these interaction patterns
to discriminative discrepancy signals. The nine tokens $\{\mathbf{v},\mathbf{a},\mathbf{t},
\mathbf{d}_{va},\mathbf{d}_{vt},\mathbf{d}_{at},\mathbf{h}_{va},\mathbf{h}_{vt},\mathbf{h}_{at}\}$
are processed by a 2-layer Transformer (8 heads), flattened to $(B,9D)$, and
classified by an MLP (LayerNorm$\to$Linear(512)$\to$GELU$\to$Dropout$\to$Linear(1)),
yielding $\ell_{\mathrm{full}}$.

\subsection{FiLM Modulation and Text-Guided Fusion}

\textbf{FiLM modulation.} The text embedding conditions video and audio before
discrepancy computation via Feature-wise Linear Modulation~\cite{perez2018film}:

\begin{equation}
  [\gamma_v; \beta_v] = \mathrm{Linear}^{\mathrm{film}}_v(\mathbf{t}), \qquad
  \mathbf{v}' = \gamma_v \odot \mathbf{v} + \beta_v,
  \label{eq:film}
\end{equation}

with an analogous transform applied to $\mathbf{a}$. This allows confident
linguistic signals to down-weight noisy visual and acoustic dimensions.

\textbf{Text-guided late fusion.} A text-only auxiliary
head~\cite{liang2024auxiliary} is trained jointly with the full-fusion path:

\begin{equation}
  \ell_{\mathrm{text}} = \mathrm{MLP}_{\mathrm{text}}(\mathbf{t}), \qquad
  \mathcal{L} = (1-w)\,\mathcal{L}_{\mathrm{BCE}}(\ell_{\mathrm{full}}, \tilde{y})
               + w\,\mathcal{L}_{\mathrm{BCE}}(\ell_{\mathrm{text}}, \tilde{y}),
  \label{eq:loss}
\end{equation}

with $w=0.5$ and optional label smoothing $\tilde{y}$. At inference the two
branches are blended:

\begin{equation}
  p = \alpha\cdot\sigma(\ell_{\mathrm{text}}) + (1-\alpha)\cdot\sigma(\ell_{\mathrm{full}}),
  \label{eq:blend}
\end{equation}

where $\alpha=0.6$ is tuned on the validation set.

\subsection{LoRA Fine-Tuning}

LoRA~\cite{hu2022lora} injects trainable low-rank matrices into query and value
projections of each encoder:

\begin{equation}
  \mathbf{W} \leftarrow \mathbf{W} + \mathbf{B}\mathbf{A},
  \label{eq:lora}
\end{equation}

with rank $r=8$ and scaling $\alpha=16$. Encoder weights remain frozen; only the
adapters and task heads are updated, substantially reducing trainable parameters
and overfitting.

\section{Experiments}
\label{sec:experiments}

\subsection{Dataset and Setup}

The BAH dataset (\Cref{tab:dataset}) provides 1,427 videos with participant-level
splits; 151 additional unlabelled videos serve as the private test set. All
experiments use a single RTX 4090 (24GB), implemented in
PyTorch~\cite{paszke2019pytorch} with HuggingFace
Transformers~\cite{wolf2020transformers}. LoRA uses PEFT~\cite{mangrulkar2022peft}.

Training uses AdamW ($\beta_1=0.9,\;\beta_2=0.999$, weight decay $10^{-2}$),
learning rate $3\times10^{-5}$ with cosine annealing to $3\times10^{-7}$,
mixed precision (fp16), and effective batch size 16. Enhancement modules applied
optionally include focal loss~\cite{lin2017focal} ($\gamma=2.0$),
CutMix~\cite{yun2019cutmix} (probability 0.5), 5-epoch LR warmup, and
multi-window training ($K=3$ uniformly-spaced windows with mean pooling). At inference, $N=5$ windows are averaged. Training stops after 15 epochs without
validation improvement; thresholds in $[0.25,0.75]$ are swept per epoch. Training converges within 20--25 minutes (roughly 50\,s/epoch).

\begin{table}[tb]
  \caption{BAH dataset splits for the ABAW11 A/H Challenge.}
  \label{tab:dataset}
  \centering
  \begin{tabular*}{\textwidth}{@{}l @{\extracolsep{\fill}} r r r @{}}
    \toprule
    Split            & Videos & A/H           & No-A/H \\
    \midrule
    Train            & 778    & 385 (49\%)    & 393 (51\%) \\
    Validation       & 124    & 75  (60\%)    & 49  (40\%) \\
    Test (labelled)  & 525    & 318 (61\%)    & 207 (39\%) \\
    Test (unlabelled)& 151    & \multicolumn{2}{c}{--} \\
    \bottomrule
  \end{tabular*}
\end{table}

\subsection{Main Results}

MDT (9-token discrepancy, FiLM, LoRA, with focal loss, CutMix, warmup, and
multi-window training) achieves 0.7408 Macro F1 on the labelled test
split---surpassing the strongest BAH baseline (0.634) by 10.7 points and
CA-AH (0.715) by 2.6 points (\Cref{tab:main}). On the private test set it
achieves 0.7368.

\begin{table}[tb]
  \caption{A/H recognition performance on the BAH labelled test set (525 videos).
    Published baselines are from Gonz\'{a}lez~\etal~\cite{gonzalez2026bah};
    CA-AH refers to the conflict-aware architecture of
    Bekhouche~\etal~\cite{bekhouche2026conflictaware}. MDT denotes our full model.}
  \label{tab:main}
  \centering
  \begin{tabular*}{0.6\textwidth}{@{}l @{\extracolsep{\fill}} r @{}}
    \toprule
    Model & Macro F1 \\
    \midrule
    BAH: ZF M-LLM (vision only)~\cite{gonzalez2026bah}  & 0.283 \\
    BAH: Video-FocalNet~\cite{gonzalez2026bah}           & 0.566 \\
    BAH: LFAN (V+A+T)~\cite{gonzalez2026bah}             & 0.593 \\
    BAH: ZF M-LLM + transcript~\cite{gonzalez2026bah}   & 0.634 \\
    CA-AH~\cite{bekhouche2026conflictaware}              & 0.715 \\
    \midrule
    \textbf{MDT (labelled test)}                         & \textbf{0.7408} \\
    \textbf{MDT (private test, 151 unlabelled)}           & \textbf{0.7368} \\
    \bottomrule
  \end{tabular*}
\end{table}

\subsection{Ablation Studies}
\label{sec:ablation}

Unless stated otherwise, ablations use the 6-token abs baseline (CA-AH
configuration: unfrozen top-2 encoder layers, 6-token absolute difference,
$\alpha=0.6$) on the labelled test split.

\subsubsection{Discrepancy Representation and Text Blend}

\Cref{tab:component} evaluates core fusion components. Removing all discrepancy
features (fusing only $[\mathbf{v};\mathbf{a};\mathbf{t}]$) achieves 0.7325,
slightly above CA-AH (0.7219), indicating that 6-token abs features alone do
not provide net benefit at this data scale---the additional tokens introduce noise
outweighing their discriminative value. Augmenting with Hadamard-product features
(9 tokens) yields 0.7322 while improving per-class calibration (F1-NoAH from
0.6457 to 0.6700). The frozen-encoder variant (all encoders frozen) reaches
0.7181.

The text blend coefficient significantly affects performance: $\alpha=0.0$
(full-fusion only) drops to 0.7080, while $\alpha=1.0$ (text-only inference)
yields 0.7185 but with the largest class-performance gap (F1-AH 0.8075\,vs.\
F1-NoAH 0.6295). The blended $\alpha=0.6$ provides the best balance at 0.7219.

\begin{table}[tb]
  \caption{Component ablation on the labelled test set. CA-AH denotes the
    6-token absolute-difference baseline~\cite{bekhouche2026conflictaware}.}
  \label{tab:component}
  \centering
  \begin{tabular*}{\textwidth}{@{}l @{\extracolsep{\fill}} r r r @{}}
    \toprule
    Configuration & Macro F1 & F1-AH & F1-NoAH \\
    \midrule
    Frozen CA-AH                                   & 0.7181 & 0.7793 & 0.6569 \\
    CA-AH (unfreeze=2, 6-token abs, $\alpha$=0.6)  & 0.7219 & 0.7982 & 0.6457 \\
    \quad $-$ All discrepancy (no conflict)         & 0.7325 & 0.8018 & 0.6632 \\
    \quad $+$ Hadamard discrepancy (9-token both)   & 0.7322 & 0.7944 & 0.6700 \\
    \quad $\alpha=0.0$ (full-fusion only)           & 0.7080 & 0.7759 & 0.6402 \\
    \quad $\alpha=1.0$ (text-only inference)        & 0.7185 & 0.8075 & 0.6295 \\
    \bottomrule
  \end{tabular*}
\end{table}

\subsubsection{Modality Analysis}

\Cref{tab:modality} isolates modality contributions using the CA-AH checkpoint.
Text alone reaches 0.7059, far above video (0.3772) or audio (0.3772). Both
video- and audio-only models collapse to predicting every sample as A/H-positive
(zero F1-NoAH), confirming that only the linguistic channel can independently
discriminate the negative class. Multimodal fusion (0.7219) improves F1-AH from
0.7608 to 0.7982 over text alone.

\begin{table}[tb]
  \caption{Single-modality and multimodal performance on the BAH labelled test set.
    All rows use the CA-AH checkpoint, with active modalities restricted to the
    indicated subsets at inference.}
  \label{tab:modality}
  \centering
  \begin{tabular*}{\textwidth}{@{}l @{\extracolsep{\fill}} r r r @{}}
    \toprule
    Active Modalities & Macro F1 & F1-AH & F1-NoAH \\
    \midrule
    Video only        & 0.3772 & 0.7544 & 0.0000 \\
    Audio only        & 0.3772 & 0.7544 & 0.0000 \\
    Text only         & 0.7059 & 0.7608 & 0.6511 \\
    Video + Audio + Text (CA-AH) & 0.7219 & 0.7982 & 0.6457 \\
    \bottomrule
  \end{tabular*}
\end{table}

\subsubsection{Training Strategy Analysis}

\Cref{tab:training} incrementally adds each module to the CA-AH baseline.
LoRA yields the largest individual gain at 0.7366 (+1.47), confirming the
superiority of parameter-efficient adaptation over layer unfreezing at this
data scale. Hadamard discrepancy (9-token) achieves 0.7322 with improved
F1-NoAH. Among enhancement modules, multi-window training is most impactful
(0.7354), suggesting broader temporal coverage benefits sparse A/H cues.
MDT combining all components achieves 0.7408.

\begin{table}[tb]
  \caption{Incremental training strategy ablation. Each row adds one component
    to the CA-AH baseline (6-token abs, top-2 unfrozen, $\alpha$=0.6).}
  \label{tab:training}
  \centering
  \begin{tabular*}{\textwidth}{@{}l @{\extracolsep{\fill}} r r r @{}}
    \toprule
    Configuration & Macro F1 & F1-AH & F1-NoAH \\
    \midrule
    CA-AH baseline                                & 0.7219 & 0.7982 & 0.6457 \\
    \quad $+$ LoRA fine-tuning                     & 0.7366 & 0.8119 & 0.6613 \\
    \quad $+$ Hadamard discrepancy (9-token)       & 0.7322 & 0.7944 & 0.6700 \\
    \quad $+$ Focal loss ($\gamma=2.0$)            & 0.7275 & 0.7836 & 0.6715 \\
    \quad $+$ CutMix augmentation                  & 0.7300 & 0.8077 & 0.6524 \\
    \quad $+$ LR warmup (5 epochs)                 & 0.7242 & 0.7883 & 0.6600 \\
    \quad $+$ Multi-window training ($K=3$)        & 0.7354 & 0.8024 & 0.6684 \\
    \quad \textbf{MDT (all combined)}              & \textbf{0.7408} & \textbf{0.7857} & \textbf{0.6959} \\
    \bottomrule
  \end{tabular*}
\end{table}

\subsection{Discussion}

\textbf{Why does removing discrepancy help?} The CA-AH baseline (6-token abs,
0.7219) underperforms the no-discrepancy variant (0.7325). On 778 training
examples, six tokens provide more parameters for the Transformer to overfit
to spurious disagreement patterns than three. The 9-token design recovers
this gap while improving class calibration (F1-NoAH 0.6700), indicating that
richer discrepancy representations distinguish genuine conflict from noise.

\textbf{Text bias correction.} Text-only inference ($\alpha=1.0$) produces a
0.178 gap between F1-AH (0.8075) and F1-NoAH (0.6295); MDT narrows this to
0.090 (0.7857 vs.\ 0.6959). The discrepancy features and FiLM modulation
provide cross-modal consistency checks that anchor negative predictions, a
mechanism unavailable to text-only approaches.

\textbf{Overfitting governs performance.} LoRA achieves the best single-module gain (0.7366) with
fewer parameters than layer unfreezing, directly supporting regularisation---not
capacity---as the limiting factor. Limitations include the 16-frame window
covering only part of each video, geometric rather than semantic discrepancy
measurement (no contrastive pre-training), and the moderate absolute performance
level reflecting the inherent difficulty of the task at this data scale.

\section{Conclusion}
\label{sec:conclusion}

MDT enriches conflict-aware fusion with a 9-token discrepancy representation
combining absolute-difference and Hadamard-product features, FiLM-based
cross-modal modulation, and LoRA fine-tuning. It achieves 0.7408 Macro F1 on
the BAH labelled split and 0.7368 on the ABAW11 leaderboard, outperforming
all published baselines and advancing the state of the art for video-level
A/H recognition. Key findings include that simple discrepancy features alone
are insufficient at this data scale---the representation must be sufficiently
rich to offset the overfitting cost of additional tokens---and that
parameter-efficient fine-tuning and temporal coverage are the most impactful
enhancement strategies.

A promising direction for future investigation is
leveraging the BAH dataset's frame-level annotations to extend the discrepancy
paradigm from video-level classification to fine-grained temporal grounding,
enabling models not only to detect whether A/H occurs in a video but to
localise when and through which modalities it manifests.

\vspace{4pt}
\noindent\textbf{Acknowledgements.} This work is supported by the Strategic
Priority Research Program of Chinese Academy of Sciences, Grant
No.~XDB0500103, and the National Natural Science Foundation of China (NSFC)
under Grant No.~62276259.

\bibliographystyle{splncs04}
\bibliography{main}
\end{document}